Preprint

# Dynamics of collective creativity in AI art competitions

**Mason Youngblood[1,*], Jeff Nusz[2] and Joel Simon[2]**

[1]Institute for Advanced Computational Science, Stony Brook University, NY, USA
[2]Morphogen, CA, USA
[*]Corresponding author: masonyoungblood@gmail.com

## ABSTRACT

**Creativity is a fundamental aspect of how culture evolves, yet the mechanisms by which groups produce novelty are notoriously difficult to infer from the historical record. Iterated learning experiments have shown that cultural transmission reliably distorts artifacts toward the inductive biases of learners, but most of this work uses linear chains between human participants, leaving open how these dynamics play out in the networked, human-AI systems that increasingly shape cultural production. In this study, we leverage one such system, Artbreeder, which hosts daily "remix parties" where users iteratively build on each other's work from a single seed image, producing branching lineages of human-AI co-created images. We analyze a dataset of 130,882 images from 368 remix parties over 13 months and find that images become simpler and converge toward common thematic "attractors" (e.g., steampunk scenes, alien architecture). We also find that while more novel "parent" images produce more novel and complex "children" that attract more likes, users paradoxically prefer to remix images that are less novel and complex. Finally, larger remix parties produce more novelty at the cost of lower complexity.**

## INTRODUCTION

Creativity is a fundamental aspect of how culture evolves because it generates the variation that other processes select for or against. However, the mechanisms by which groups produce novelty are notoriously difficult to infer from the historical record. As a result, creativity is often discussed but rarely measured[1,2]. In cultural evolution and cognitive science, people think about creativity as a search process that results in something that is "novel, surprising, and valuable"[3]. While value is domain-specific and difficult to measure, novelty is a far more tractable proxy for creative exploration. Here, we operationalize creativity as the generation of novelty—variants that exist in a less probable area of representational space[4].

Culture is a collective phenomenon, where individuals iteratively build on the ideas and behaviors of others in their group[5], typically by recombining and modifying existing artifacts[6]. A central paradigm for studying these cumulative dynamics is iterated learning, where content is passed along transmission chains and progressively shaped by learners' biases[7]. Iterated learning reliably produces simpler, more structured outputs that converge toward attractors shaped by cognitive and production biases, as shown in examples ranging from artificial languages[7] to writing systems[8] to songs[9]. Recent work has extended the paradigm in two directions: from linear chains to networks of interacting participants[10], and from purely human transmission to hybrid human-AI systems in which content is jointly created by people and generative models[11,12]. Digital creative platforms where users build on one another's work with generative AI are naturally-occurring instances of this more complex setting, and give us unprecedented access to how cumulative culture unfolds when the transmission graph branches and the learners are humans collaborating with AI.

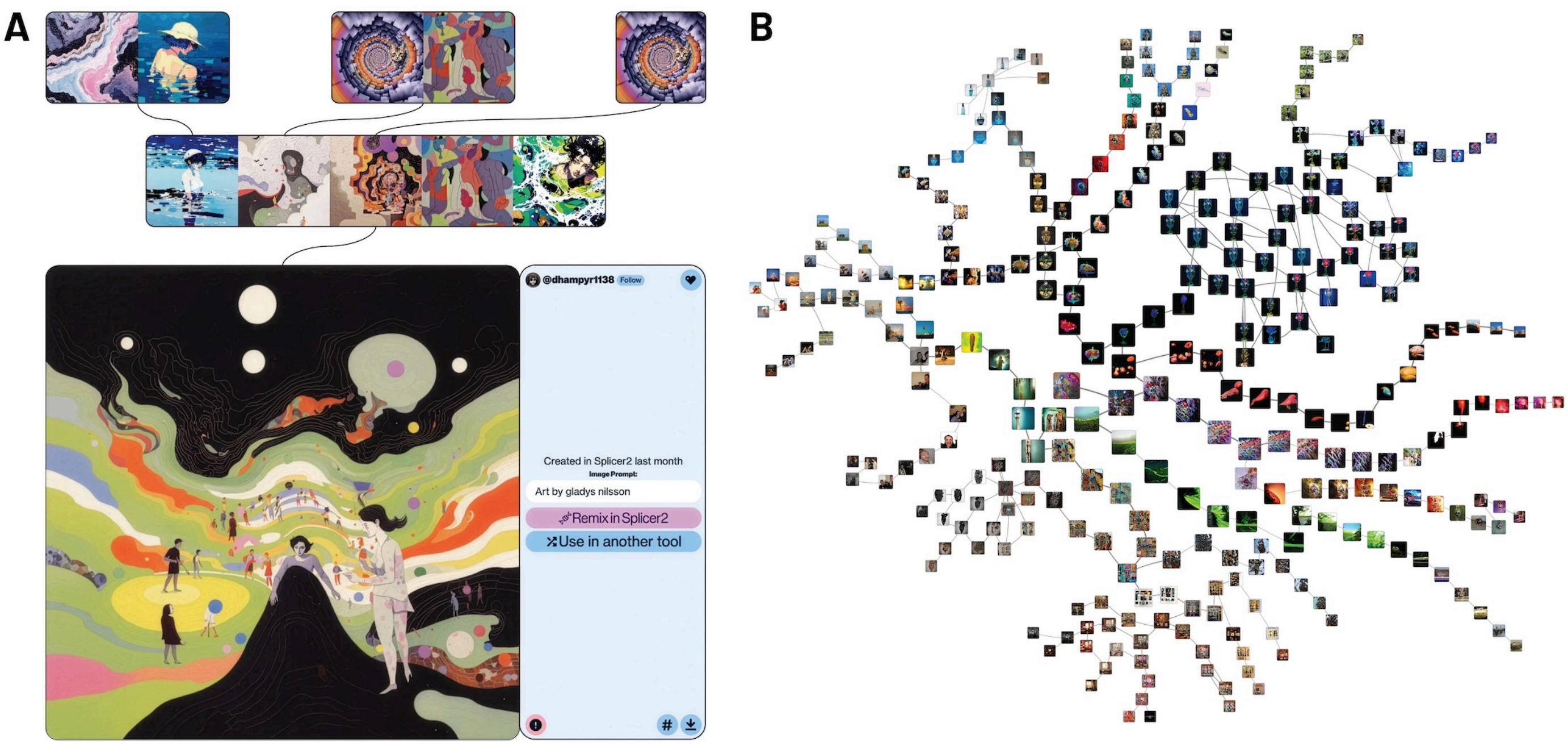

**Figure 1:** On the left (panel A)—an example of an entry in an Artbreeder remix party, with the parents and grandparents above, and the text prompt to the right. On the right (panel B)—a print from Joel Simon's *Latent Lineages* series shown at Art Basel in 2025, depicting the branching lineages of remixing from a single image on Artbreeder.

Artbreeder is a collaborative, AI visual art platform, originally based on BigGAN and StyleGAN[13,14] and now using Stable Diffusion and FLUX[15,16]. These deep generative models take user-supplied images and text prompts as inputs and produce new images as outputs, allowing users to mix and modify existing content rather than always generating from scratch. Unlike other major platforms, like Midjourney, Artbreeder emphasizes this kind of flexible modification and remixing[17]. All recent Artbreeder tools allow users to iteratively combine multiple images and text prompts, adjusting the relative weights assigned to each input to shape the output. The four most heavily used of these tools —Collage, Composer, Mixer, and Splicer2—differ in the interface used to combine inputs, but all share the same basic structure of remixing one or more parent images and an optional text prompt into a single child image.

Since mid-April 2024, Artbreeder has been hosting daily "remix parties" where the platform posts a seed image as the first entry, and users respond to the following prompt: "(1) Remix ANY image that anyone has entered into this party. (2) When you make a remix, try to keep some aspect of the original while adding your own twist to take things in a new direction. (3) When you make something you like, share what you've done by entering it here." What follows are lineages of remixes that branch out from the seed (Figure 1). During each remix party, users have access to a timeline of others' images, and if they click on an image they can access the underlying generative process (e.g., "user487 combined image351 and image713 with prompt316 to produce image906"). Users like and comment on each image, which ends up being the measure of "success" in each remix party.

In this study, we analyze a large corpus of remix parties from Artbreeder as a naturally-occurring, networked, human-AI iterated learning system. Classical iterated learning predicts that lineages should simplify and converge toward learner-biased attractors over the course of transmission, and these predictions can be tested directly here. The networked and hybrid nature of Artbreeder also raises questions that linear, human-only chains cannot address—for example, whether drawing on multiple, diverse parents injects novelty, and how individual choice of which image to remix interacts with collective outcomes. We frame the analysis around three questions: (1) how the properties of parent images and text prompts shape the novelty and complexity of child images, (2) how the properties of child images predict downstream success in the form of likes and future remixes, and (3) how population size, depth, and other party-level factors affect creative production. To do this, we project all images and text prompts into a shared embedding space using OpenCLIP[18], calculate complexity and novelty using image segmentation and density estimation[19,20], and use Bayesian structural equation modeling to estimate the relationships between variables[21].

## METHODS

### Data

We collected all data from all remix parties that occurred between April 10, 2024 and May 5, 2025. Images from parties that lacked the original seed images, and images with missing data (e.g., incomplete "key" in the database, missing link to the seed image), were removed from the analysis. Additionally, we only analyzed remixes that were produced with the four most heavily used tools on Artbreeder that allow users to combine multiple inputs: Collage, Composer, Mixer, and Splicer2. After applying these filters, we were left with 130,882 images from 368 remix parties (~77.5% of the original dataset).

Note that in order to compute the predicted probability of images in an unbiased way, we needed to create a density estimation model that was trained on different data than it is applied to. To do this, we split the data up into odd and even remix parties, used the odd parties to train the model, and only analyzed outcomes in the even parties (see Novelty below).

**Table 1:** The variables included in the causal modeling. Note that the missing data here correspond to cases where there was only a single parent image, or where a text prompt was not used.

| Category | Variable | Missing data | Transform | Model family |
|---|---|---|---|---|
| Parent image | Distance (b/t parent images) | ~37.2% | z-scale | |
| Parent image | Complexity (# segments) | | log and z-scale | |
| Parent image | Novelty (inv. probability) | | log and z-scale | |
| Parent image | # parents | | z-scale | |
| Parent text prompt | Distance (from parent images) | ~32.1% | z-scale | |
| Parent text prompt | Novelty (inv. probability) | ~32.1% | log and z-scale | |
| Parent text prompt | Length (# characters) | ~32.1% | z-scale | |
| Child image | Distance from parents | | z-scale | Gaussian |
| Child image | Complexity (# segments) | | log and z-scale | Gaussian |
| Child image | Novelty (inv. probability) | | log and z-scale | Gaussian |
| Outcome | # grandchildren | | | Poisson |
| Outcome | # likes | | | Poisson |
| Remix party | Depth (of child image) | | z-scale | |
| Remix party | Population size | | z-scale | |
| Remix party | Time | | z-scale | |

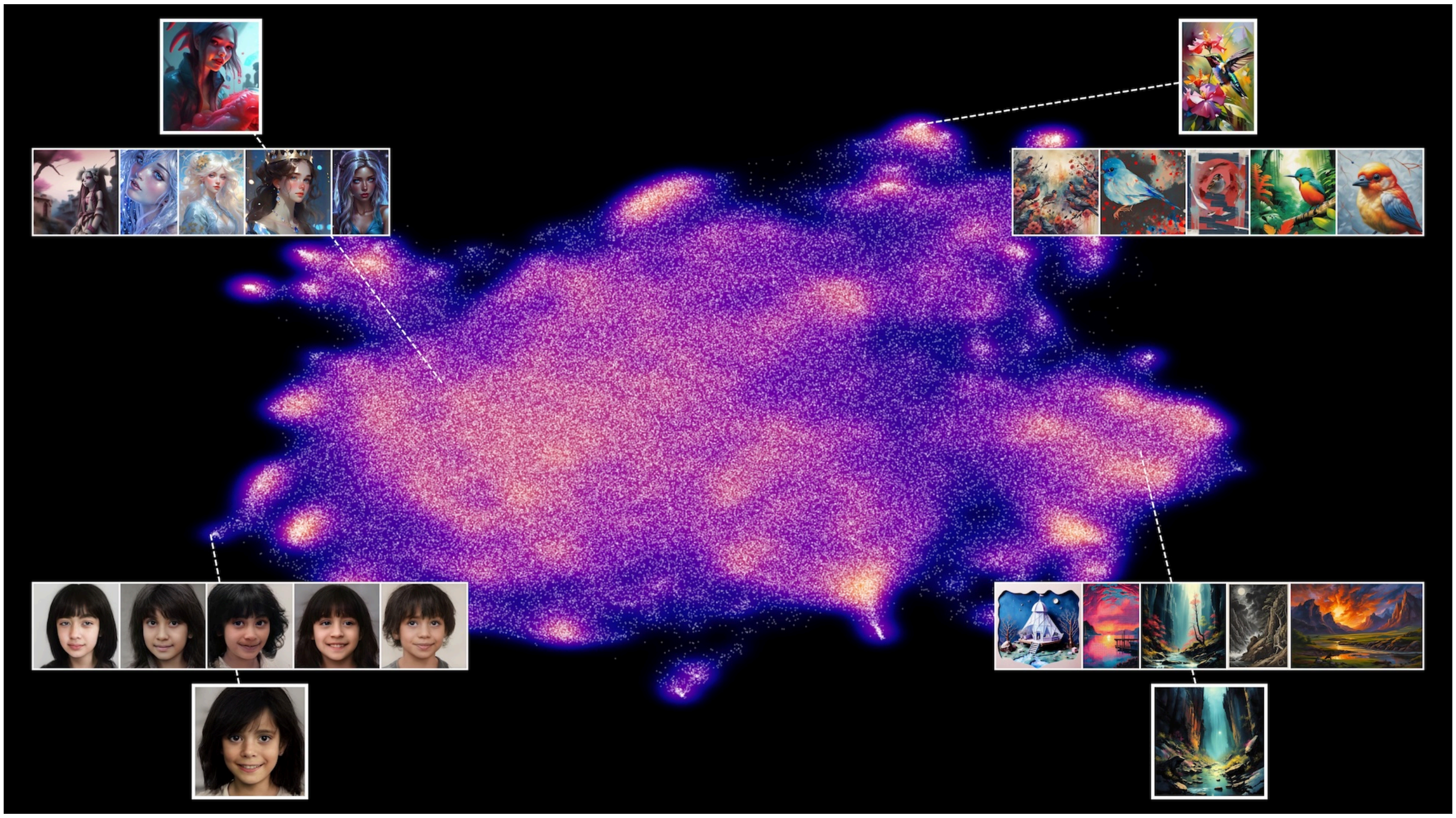

**Figure 2:** A UMAP projection of the OpenCLIP embedding space, with kernel density estimation underneath. This is just for visualization purposes, as the real analysis occurs in the raw embedding space and with neural density estimation. The four large images are random focal images, and the five smaller images are their nearest neighbors in embedding space.

Images on Artbreeder are in the public domain under a Creative Commons Zero license, and all data were collected with the consent of the Artbreeder developers.

### Embeddings

Embeddings for images and text prompts were extracted using OpenCLIP[18], an open-source version of OpenAI's CLIP model that was built to project both images and text into a shared embedding space. Specifically, we used the CLIP-ViT-H-14-laion2B-s32B-b79K checkpoint (https://huggingface.co/laion/CLIP-ViT-H-14-laion2B-s32B-b79K). The output of OpenCLIP is a 1024-dimensional vector for each image and text prompt.

### Distance

Distances between images and text prompts were computed using the cosine distance (1 - cosine similarity) applied to the embeddings.

### Novelty

To compute the novelty of images and text prompts, we conducted density estimation using the *denmarf* package in Python[20]. A masked autoregressive flow, with the default architecture in *denmarf*, was trained on the embeddings of images and text prompts from the odd remix parties, and used to predict the probability of images and text prompts from the even remix parties. Novelty was operationalized as the inverse of predicted probability. We emphasize that this measure captures statistical atypicality in representational space—an image (or prompt) in a low-density region is unusual relative to the corpus, but is not necessarily aesthetically or semantically valuable.

### Complexity

To compute the complexity of images, we counted the number of segments (i.e., groups of similar pixels) using Meta's Segment Anything Model (SAM)[19], which heavily correlates with perceived image complexity[22]. Character count was used as a coarse proxy of text prompt complexity.

### Depth

In addition to the time at which an image was submitted to a remix party, we were also interested in the "depth" of images in the lineages—the number of steps between that image and the original seed image.

### Causal modeling

Our model includes 15 variables related to the parent and child images involved in these remix parties, which have complex causal dependencies between one another (see Table 1). As a result, we took a causal modeling approach and built a structural equation model in Stan using the *brms* package in R[21]. The structure of our model can be seen in Figure 3.

Besides the number of likes and grandchildren, all other variables were z-scaled prior to modeling, and those with lognormal distributions were also log-transformed (Table 1). Seed images were excluded as children in the model, but were included in the parent image measures. Some variables had missing data, due to the fact that some Artbreeder tools do not require multiple parent images or the use of prompts. Missing data (Table 1) were handled with mean imputation, as multiple imputation was prohibitively computationally expensive given the complexity of the model. Because mean imputa-tion assumes missingness is unrelated to the underlying values and shrinks the imputed variable toward its mean, it likely attenuates associations involving text prompts and between-parent distance, and underestimates uncertainty in those coefficients. The reported effects on these variables should therefore be read as conservative lower bounds.

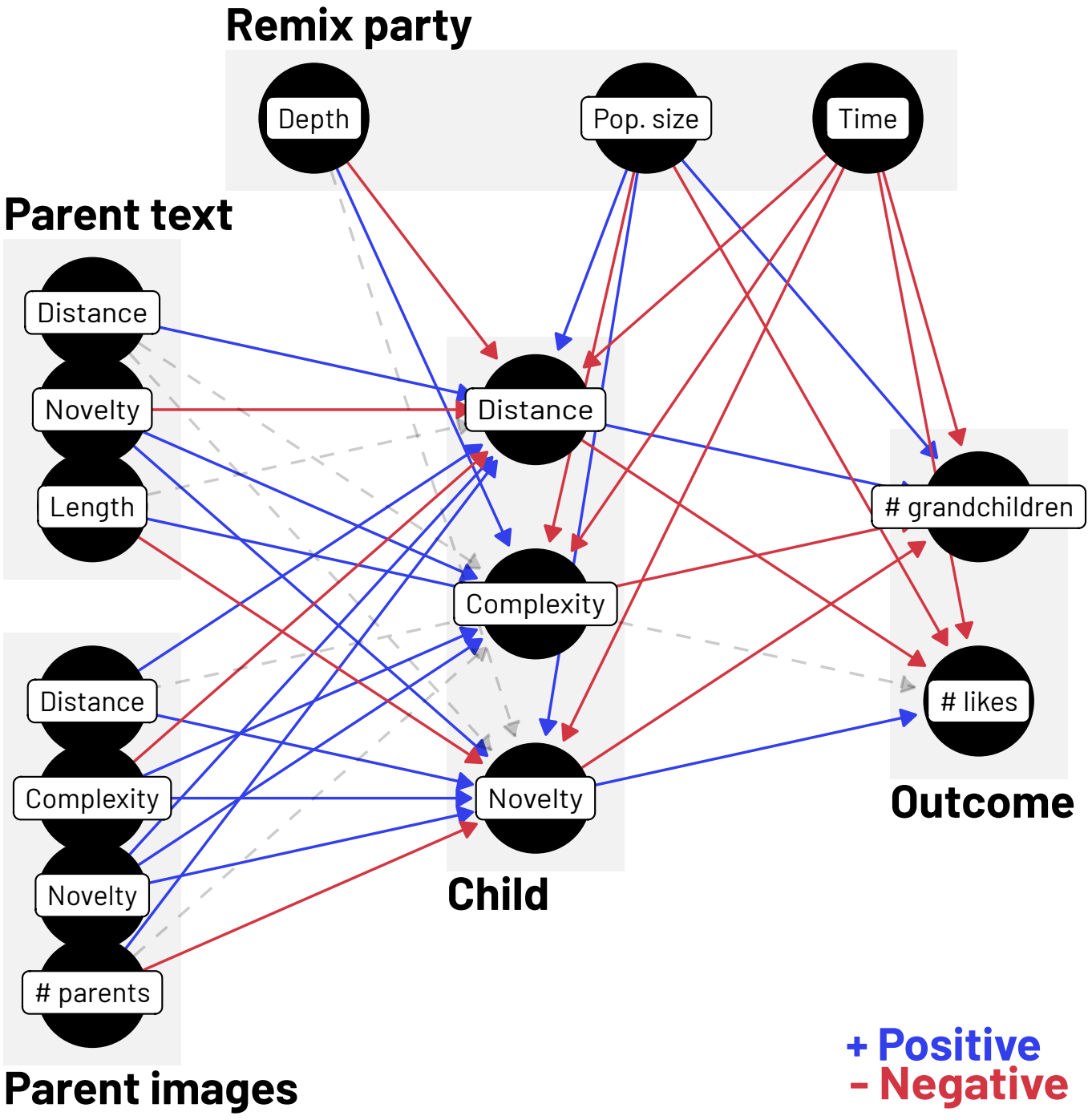

**Figure 3:** A directed acyclic graph that shows the results of the Bayesian structural equation model. Each node is a variable, and each arrow is a causal relationship. Dotted grey is no relationship, blue is a positive relationship, and red is a negative relationship. Relationships are determined to be positive or negative if the two bounds of the 95% credible interval do not overlap zero. Different categories of variables are labelled with gray boxes (e.g., all measures of text prompts are in the top left).

Gaussian priors with mean of 0 and standard deviation of 1 were used for all model effects, and Gaussian priors with mean of 0 and standard deviation of 0.5 were used for all variances. The three child image measures (distance, novelty, and complexity) are continuous after log- and z-transformation and were modeled as Gaussian distributions, while the two outcome variables (number of likes and number of grandchildren) are non-negative counts and were modeled as Poisson distributions on their raw counts (Table 1). The model was run across 15 chains of 2,000 iterations each.

## RESULTS

The results of the Bayesian structural equation model can be seen in Figure 3 and Table 2. The web of causal relationships is fairly complicated, so we will focus on what we think are the five most interesting results related to collective creativity:

1. More novel, complex, and diverse (i.e., far apart from one another) parent images lead to more novel and more complex child images.
2. More novel child images get more likes.
3. More novel child images have fewer grandchildren, or are less likely to be remixed.
4. Images become less novel, less complex, and closer together as time and depth increase. A qualitative exploration of the embedding suggests that the images converge towards common "attractors" (e.g., steampunk scenes, alien architecture).
5. Remix parties with larger populations (i.e., more participants) produce images that are more novel but less complex.

The fitted model accounts for a substantial portion of the variance in complexity ($R^2 = 0.454$), novelty ($R^2 = 0.495$), distance ($R^2 = 0.580$), and likes ($R^2 = 0.428$), but only captures a small amount of the variance in the number of grandchildren ($R^2 = 0.090$).

## DISCUSSION

Our analysis shows that more novel and complex parent images lead to more novel and complex child images—a form of transmission fidelity that mirrors classical findings from linear iterated learning chains[7]. More strikingly, the diversity of parent images (i.e., how spread out they are in space) also boosts novelty. This effect is invisible in traditional linear chains, which transmit from a single predecessor, and highlights a distinctive feature of networked cultural production: drawing on multiple, dissimilar ancestors injects variation into lineages.

More novel child images are more likely to get likes, but less likely to be remixed in the future. Similarly, greater complexity reduces the probability of future remixes. These findings highlight an interesting tension —novel images are aesthetically appealing, but simpler inputs are preferred for remixing. Classical iterated learning typically obscures this kind of asymmetry: the "learner" and the "selector" are the same person, so there is no way to distinguish what content is valued from what content is passed on. Networked platforms like Artbreeder make this decomposition visible, and suggest that the simplification characteristic of iterated learning chains may be driven as much by producers choosing easier-to-modify inputs as by individual-level transmission biases. In hip-hop and electronic music, some of the most common "samples" are drum breaks, which are simple rhythmic sequences with no melodic elements that can be easily modified, often in quite extreme ways[23]. Something similar may happen in the visual art realm, where artists prefer to combine simpler inputs because they are easier to modify. It is important to note, though, that the model only explains a small amount of the variance in the number of grandchildren, so this result should be interpreted with caution.

We also found that images become less novel, less complex, and closer together over the course of remix parties—appearing to converge towards common "attractors" (e.g., steampunk scenes, alien architecture). This is one of the core predictions of iterated learning: when content is repeatedly transmitted, it is gradually distorted toward whatever the transmitters find easier to represent, produce, or remember, resulting in convergence[7]. The same pattern has been documented for artificial languages[7], writing systems[8], and songs[9], and it emerges here despite the branching, hybrid structure of remix parties. What makes this setting distinctive is that the "transmitter" is not just a human: biases in the training data of the underlying generative models push remixes toward certain kinds of content[12], platform norms shape what users choose to build on, and human cognitive biases operate on top of both. The persistence of classical iterated learning signatures even in this human-AI hybrid setting suggests that a broader view of the "learner"—one that encompasses both cognitive and algorithmic biases—may be necessary for understanding cultural evolution in AI-mediated environments.

**Table 2:** The estimates from the Bayesian structural equation model outlined in Figure 3. Effects with 95% credible intervals that do not overlap zero are interpreted as strong effects, and are marked with an asterisk.

| Outcome | Predictor | Est. | 2.5% | 97.5% | |
|---|---|---|---|---|---|
| Complexity (child) | # parents | 0.005 | −0.003 | 0.012 | |
| Complexity (child) | Complexity (parent image) | 0.442 | 0.436 | 0.449 | * |
| Complexity (child) | Novelty (parent image) | 0.052 | 0.046 | 0.059 | * |
| Complexity (child) | Pop. size | −0.045 | −0.061 | −0.029 | * |
| Complexity (child) | Time | −0.010 | −0.016 | −0.003 | * |
| Complexity (child) | Depth | 0.009 | 0.002 | 0.016 | * |
| Complexity (child) | Distance (b/t parent images) | −0.001 | −0.009 | 0.008 | |
| Complexity (child) | Length (parent text) | 0.075 | 0.064 | 0.086 | * |
| Complexity (child) | Novelty (parent text) | 0.042 | 0.034 | 0.051 | * |
| Complexity (child) | Distance (parent text to parent images) | 0.005 | −0.002 | 0.013 | |
| Novelty (child) | # parents | −0.130 | −0.137 | −0.122 | * |
| Novelty (child) | Complexity (parent image) | 0.057 | 0.051 | 0.063 | * |
| Novelty (child) | Novelty (parent image) | 0.457 | 0.450 | 0.463 | * |
| Novelty (child) | Pop. size | 0.022 | 0.009 | 0.035 | * |
| Novelty (child) | Time | −0.010 | −0.016 | −0.004 | * |
| Novelty (child) | Depth | −0.003 | −0.009 | 0.003 | |
| Novelty (child) | Distance (b/t parent images) | 0.030 | 0.022 | 0.038 | * |
| Novelty (child) | Length (parent text) | −0.037 | −0.048 | −0.027 | * |
| Novelty (child) | Novelty (parent text) | 0.253 | 0.245 | 0.262 | * |
| Novelty (child) | Distance (parent text to parent images) | 0.004 | −0.004 | 0.011 | |
| Distance (child) | # parents | 0.192 | 0.186 | 0.199 | * |
| Distance (child) | Complexity (parent image) | −0.007 | −0.013 | −0.002 | * |
| Distance (child) | Novelty (parent image) | 0.261 | 0.255 | 0.267 | * |
| Distance (child) | Pop. size | 0.054 | 0.041 | 0.067 | * |
| Distance (child) | Time | −0.013 | −0.019 | −0.007 | * |
| Distance (child) | Depth | −0.038 | −0.044 | −0.032 | * |
| Distance (child) | Distance (b/t parent images) | 0.347 | 0.340 | 0.354 | * |
| Distance (child) | Length (parent text) | −0.001 | −0.010 | 0.009 | |
| Distance (child) | Novelty (parent text) | −0.024 | −0.032 | −0.017 | * |
| Distance (child) | Distance (parent text to parent images) | 0.510 | 0.503 | 0.517 | * |
| # likes | Complexity (child) | 0.002 | −0.004 | 0.009 | |
| # likes | Novelty (child) | 0.040 | 0.034 | 0.047 | * |
| # likes | Distance (parent to child) | −0.011 | −0.017 | −0.004 | * |
| # likes | Pop. size | −0.241 | −0.284 | −0.202 | * |
| # likes | Time | −0.238 | −0.245 | −0.232 | * |
| # grandchildren | Complexity (child) | −0.070 | −0.079 | −0.061 | * |
| # grandchildren | Novelty (child) | −0.012 | −0.022 | −0.002 | * |
| # grandchildren | Distance (parent to child) | 0.040 | 0.031 | 0.049 | * |
| # grandchildren | Pop. size | 0.075 | 0.051 | 0.098 | * |
| # grandchildren | Time | −0.437 | −0.446 | −0.427 | * |

Finally, population size (i.e., the number of participants in a remix party) has a positive effect on novelty and a negative effect on complexity. The novelty result is broadly consistent with previous work showing that population size has a positive effect on cultural diversity and innovation[24], although the empirical literature on this is quite variable and includes some counterexamples[25]. The result for complexity is a bit more puzzling. Historically, population size has been associated with cultural complexity[26], but more recent work shows that population size can sometimes inhibit cultural complexity[27]. Some cultural forms also simplify over the course of transmission[8]. We suspect that this latter mechanism is what drives our complexity result: larger remix parties tend to have longer lineages, which gives more opportunities for the simplification biases characteristic of iterated learning to accumulate. In other words, the negative effect of population size on complexity here may be indirect, mediated by the depth of transmission rather than reflecting a direct effect of group size on what users produce.

Several limitations shape how these results should be interpreted. We lack detailed information about who participates in remix parties and why, so the patterns we observe may partly reflect platform-specific norms or technical comfort rather than general features of collective

creativity—for example, the preference for remixing less novel and less complex images could be partly strategic, with users gravitating toward thematic attractors they know will be well received, rather than a cognitive bias. Relatedly, remix party instructions explicitly ask users to "keep some aspect of the original while adding your own twist," which may push users away from selecting highly novel images to remix and dampen the novelty of the children they produce. Our measure of novelty also captures statistical atypicality in representational space, not creative value in the sense of[3], so low-density images may be genuinely innovative or simply incoherent outputs of the underlying generative models; the fact that more novel children attract more likes is consistent with the former, but a more complete test would require perceptual ratings from human participants. Relatedly, the attractors that emerge over the course of remix parties almost certainly reflect a mixture of human cognitive biases[7], biases in the training data of the underlying generative models[12], and platform-specific norms, and observational data alone cannot fully disentangle these sources.

In conclusion, users of the digital art platform Artbreeder navigate a complex landscape of creative trade-offs. Consumers prefer novelty, but producers avoid it, favoring simpler and more familiar material for remixing. The tendency of images to converge and simplify over time mirrors the classical findings of iterated learning experiments, even though transmission here is networked, driven by collective choice, and mediated by generative AI. Big data from online platforms has revolutionized our understanding of large-scale cultural evolutionary processes, but the cumulative creative processes that generate novelty—and the ways they shift when the learners are hybrid human-AI systems—remain understudied. Digital art platforms are well-positioned to fill in this gap, acting as naturally-occurring iterated learning experiments that bridge individual- and population-level cultural dynamics. As generative AI becomes a more pervasive collaborator in cultural production, understanding these hybrid creative ecosystems is no longer optional—it is essential for predicting how culture will evolve in the years to come.

## ACKNOWLEDGMENTS

We would like to thank Katie Mudd and Margaret Schedel for their valuable feedback throughout the development of this project. We would also like to thank Stony Brook Research Computing and Cyberinfrastructure, and the Institute for Advanced Computational Science at Stony Brook University for access to the SeaWulf computing system, made possible by grants from the National Science Foundation (1531492 and Major Research Instrumentation award 2215987), with matching funds from Empire State Development's Division of Science, Technology and Innovation (NYSTAR) program (contract C210148). AI tools were used to assist in the editing process for this manuscript.

## DATA AND CODE AVAILABILITY

All code for this project is available on GitHub at https://github.com/masonyoungblood/artbreeder. Data are available upon request from Artbreeder.

## REFERENCES

1. Fogarty, L., Creanza, N., and Feldman, M.W. (2015). Cultural evolutionary perspectives on creativity and human innovation. Trends in Ecology & Evolution. https://doi.org/10.1016/j.tree.2015.10.004.
2. Perry, S., Carter, A., Smolla, M., Akçay, E., Nöbel, S., Foster, J.G., and Healy, S.D. (2021). Not by transmission alone: the role of invention in cultural evolution. Philosophical Transactions of the Royal Society B. https://doi.org/10.1098/rstb.2020.0049.
3. Boden, M.A. (1998). Creativity and artificial intelligence. Artificial intelligence. https://doi.org/10.1016/S0004-3702(98)00055-1.
4. Frascaroli, J., Leder, H., Brattico, E., and Cruys, S.V. de (2023). Aesthetics and predictive processing: grounds and prospects of a fruitful encounter. Philosophical Transactions of the Royal Society B. https://doi.org/10.1098/rstb.2022.0410.
5. Falandays, J.B., Kaaronen, R.O., Moser, C., Rorot, W., Tan, J., Varma, V., Williams, T., and Youngblood, M. (2023). All intelligence is collective intelligence. Journal of Multiscale Neuroscience. https://doi.org/10.56280/1564736810.
6. Buskell, A., Enquist, M., and Jansson, F. (2019). A systems approach to cultural evolution. Palgrave Communications. https://doi.org/10.1057/s41599-019-0343-5.
7. Kirby, S., Griffiths, T., and Smith, K. (2014). Iterated Learning and the Evolution of Language. Current Opinion in Neurobiology. https://doi.org/10.1016/j.conb.2014.07.014.
8. Kelly, P., Winters, J., Miton, H., and Morin, O. (2021). The predictable evolution of letter shapes: an emergent script of West Africa recapitulates historical change in writing systems. Current Anthropology. https://doi.org/10.1086/717779.
9. Anglada-Tort, M., Harrison, P.M.C., Lee, H., and Jacoby, N. (2023). Large-scale iterated singing experiments reveal oral transmission mechanisms underlying music evolution. Current Biology. https://doi.org/10.1016/j.cub.2023.02.070.
10. Marjieh, R., Anglada-Tort, M., Griffiths, T., and Jacoby, N. (2025). Characterizing the interaction of cultural evolution mechanisms in experimental social networks. arXiv. https://doi.org/10.48550/arXiv.2502.12847.
11. Shiiku, S., Marjieh, R., Anglada-Tort, M., and Jacoby, N. (2025). The dynamics of collective creativity in human-AI hybrid societies. arXiv. https://doi.org/10.48550/arXiv.2502.17962.
12. Acerbi, A., and Stubbersfield, J. (2023). Large language models show human-like content biases in transmission chain experiments. Proceedings of the National Academy of Sciences. https://doi.org/10.1073/pnas.2313790120.
13. Brock, A., Donahue, J., and Simonyan, K. (2019). Large scale GAN training for high fidelity natural image synthesis. International Conference on Learning Representations. https://doi.org/10.48550/arXiv.1809.11096.
14. Karras, T., Laine, S., and Aila, T. (2020). A style-based generator architecture for generative adversarial networks. IEEE Transactions on Pattern Analysis and Machine Intelligence. https://doi.org/10.1109/TPAMI.2020.2970919.
15. Podell, D., English, Z., Lacey, K., Blattmann, A., Dockhorn, T., Muller, J., Penna, J., and Rombach, R. (2023). SDXL: improving latent diffusion models for high-resolution image synthesis. arXiv. https://doi.org/10.48550/arXiv.2307.01952.
16. Black Forest Labs, Batifol, S., Blattmann, A., Boesel, F., Consul, S., Diagne, C., Dockhorn, T., English, J., English, Z., Esser, P., et al. (2025). FLUX.1 Kontext: Flow Matching for In-Context Image Generation and Editing in Latent Space. arXiv. https://doi.org/10.48550/arXiv.2506.15742.

17. Zeilinger, M. (2021). Generative adversarial copy machines. Culture Machine. https://culturemachine.net/vol-20-machine-intelligences/generative-adversarial-copy-machines-martin-zeilinger/.
18. Cherti, M., Beaumont, R., Wightman, R., Wortsman, M., Ilharco, G., Gordon, C., Schuhmann, C., Schmidt, L., and Jitsev, J. (2023). Reproducible scaling laws for contrastive language-image learning. IEEE/CVF Conference on Computer Vision and Pattern Recognition. https://doi.org/10.1109/CVPR52729.2023.00276.
19. Kirillov, A., Mintun, E., Ravi, N., Mao, H., Rolland, C., Gustafson, L., Xiao, T., Whitehead, S., Berg, A.C., Lo, W.-Y., et al. (2023). Segment anything. arXiv. https://doi.org/10.48550/arXiv.2304.02643.
20. Lo, R.K.L. (2023). denmarf: a Python package for density estimation using masked autoregressive flow. arXiv. https://doi.org/10.48550/arXiv.2305.14379.
21. Bürkner, P.-C. (2017). brms: An R Package for Bayesian Multilevel Models Using Stan. Journal of Statistical Software. https://doi.org/10.18637/jss.v080.i01.
22. Shen, T., Nath, S.S., Brielmann, A., and Dayan, P. (2024). Simplicity in complexity: explaining visual complexity using deep segmentation models. ICLR Workshop on Representational Alignment. https://doi.org/10.48550/arXiv.2403.03134.
23. Youngblood, M. (2019). Conformity bias in the cultural transmission of music sampling traditions. Royal Society Open Science. https://doi.org/10.1098/rsos.191149.
24. Deffner, D., Kandler, A., and Fogarty, L. (2022). Effective population size for culturally evolving traits. PLOS Computational Biology. https://doi.org/10.1371/journal.pcbi.1009430.
25. Youngblood, M., and Passmore, S. (2026). Simulation-based inference with deep learning suggests speed climbers combine innovation and copying to improve performance. Proceedings of the Royal Society B. https://doi.org/10.1098/rspb.2025.1433.
26. Kline, M., and Boyd, R. (2010). Population size predicts technology complexity in Oceania. Proceedings of the Royal Society B. https://doi.org/10.1098/rspb.2010.0452.
27. Fay, N., De Kleine, N., Walker, B., and Caldwell, C. (2019). Increasing population size can inhibit cumulative cultural evolution. Proceedings of the National Academy of Sciences. https://doi.org/10.1073/pnas.1811413116.